\documentclass{article} 
\usepackage{iclr2020_conference,times}
\usepackage{subfiles}
\usepackage[pdftex]{graphicx}

\usepackage{amsmath,amsfonts,bm}

\def\eqref#1{equation~\ref{#1}}

\def\1{\bm{1}}

\DeclareMathAlphabet{\mathsfit}{\encodingdefault}{\sfdefault}{m}{sl}
\SetMathAlphabet{\mathsfit}{bold}{\encodingdefault}{\sfdefault}{bx}{n}

\usepackage[utf8]{inputenc}
\usepackage{hyperref}
\usepackage{url}

\title{Bridging the ELBO and MMD}

\author{Talip Uçar \\
Department of Computer Science\\
University College London\\
\texttt{ucabtuc@ucl.ac.uk} \\
}

\begin{document}

\maketitle

\begin{abstract}
One of the challenges in training generative models such as the variational auto encoder (VAE) is avoiding posterior collapse. When the generator has too much capacity, it is prone to ignoring latent code. This problem is exacerbated when the dataset is small, and the latent dimension is high. The root of the problem is the ELBO objective, specifically the Kullback–Leibler (KL) divergence term in objective function \citep{zhao2019infovae}. This paper proposes a new  objective function to replace the KL term with one that emulates the maximum mean discrepancy (MMD) objective. It also introduces a new technique, named latent clipping, that is used to control distance between samples in latent space. A probabilistic autoencoder model, named $\mu$-VAE, is designed and trained on MNIST and MNIST Fashion datasets, using the new objective function and is shown to outperform models trained with ELBO and $\beta$-VAE objective. The $\mu$-VAE is less prone to posterior collapse, and can generate reconstructions and new samples in good quality. Latent representations learned by $\mu$-VAE are shown to be good and can be used for downstream tasks such as classification.  

\end{abstract}

\section{Introduction}

Autoencoders(AEs) are used to learn low-dimensional representation of data. They can be turned into generative models by using adversarial, or variational training. In the adversarial approach, one can directly shape the posterior distribution over the latent variables by either using an additional network called a Discriminator \citep{makhzani2015adversarial}, or using the encoder itself as a discriminator \citep{huang2018introvae}. AEs trained with variational methods are called Variational Autoencoders (VAEs) \citep{kingma2014adam, rezende2014stochastic}. Their objective maximizes the variational lower bound (or evidence lower bound, ELBO) of $p_{\theta}(x)$. Similar to AEs, VAEs contain two networks: 
    
\paragraph{Encoder - Approximate inference network:} In the context of VAEs, the encoder is a recognition model $q_{\phi}(z|x)$\footnote{$\phi$ refers to parameters of encoder while $\theta$ is parameters of decoder.}, which is an approximation to the true posterior distribution over the latent variables, $p_\theta(z|x)$. The encoder tries to map high-level representations of the input $x$ onto latent variables such that the salient features of $x$ are encoded on $z$.
    
\paragraph{ Decoder - Generative network:} The decoder learns a conditional distribution $p_{\theta}(x|z)$ and has two tasks: i) For the task of reconstruction of input, it solves an inverse problem by taking mapped latent z computed using output of encoder and predicts what the original input is (i.e. reconstruction $x'\approx x$). ii) For generation of new data, it samples new data $x'$, given the latent variables z. 
    
During training, encoder learns to map the data distribution $p_{d}(x)$ to a simple distribution such as Gaussian while the decoder learns to map it back to data distribution $p(x)$ \footnote{Note that output distribution of decoder is model distribution $p(x)$, not data distribution $p_{d}(x)$}. VAE's objective function has two terms: log-likelihood term (reconstruction term of AE objective function) and a prior regularization term \footnote{One such term would be Kullback-Leibler (KL) divergence, which is a measure of how similar two probability distributions are.}. Hence, VAEs add an extra term to AE objective function, and approximately maximizes the log-likelihood of the data, $\log p(x)$, by maximizing the evidence lower bound (ELBO):

        \begin{equation}
        \mathcal{L}_{\mathrm{ELBO}} = \mathbb{E}_{p_{d}(x)}\left[\mathbb{E}_{q_\phi(z \vert x)} \left[\log p_\theta(x \vert z) \right]\right] -\mathbb{E}_{p_{d}(x)}\left[\mathrm{KL}(q_\phi(z\vert x) \Vert p(z))\right]  
    \end{equation}

Maximizing ELBO does two things:
\begin{itemize}
    \itemsep0em 
    \item Increase the probability of generating each observed data x.
    \item Decrease distance between estimated posterior $q(z|x)$ and prior distribution $p(z)$, pushing KL term to zero. Smaller KL term leads to less informative latent variable.
\end{itemize}

Pushing KL terms to zero encourages the model to ignore latent variable. This is especially true when the decoder has a high capacity. This leads to a phenomenon called posterior collapse in literature \citep{razavi2019preventing, chen2016variational, dieng2018avoiding, kim2018semi,van2017neural,bowman2015generating, kingma2016improved, sonderby2016train, zhao2017towards}.

This work proposes a new method to mitigate posterior collapse. The main idea is to modify the KL term of the ELBO such that it emulates the MMD objective \citep{gretton2007kernel, zhao2019infovae}. In ELBO objective, minimizing KL divergence term pushes mean and variance parameters of each sample at the output of encoder towards zero and one respectively. This , in turn, brings samples closer, making them indistinguishable. The proposed method replaces the KL term in the ELBO in order to encourage samples from latent variable to spread out while keeping the aggregate mean of samples close to zero. This enables the model to learn a latent representation that is amenable to clustering samples which are similar. As shown in later sections, the proposed method enables learning good generative models as well as good representations of data. The details of the proposal are discussed in Section~\ref{proposal_uvae}.

\section{Related Work}
In the last few years, there have been multiple proposals on how to mitigate posterior collapse. These proposals are concentrated around i) modifying the ELBO objective, ii) imposing a constraint on the VAE architecture, iii) using complex priors, iv) changing distributions used for the prior and the posterior v) or some combinations of these approaches. Modifications of the ELBO objective can be done through annealing the KL term \citep{sonderby2016train, bowman2015generating}, lower-bounding the KL term to prevent it from getting pushed to zero \citep{razavi2019preventing}, controlling KL capacity by upper bounding it to a pre-determined value \citep{burgess2018understanding} or lower-bounding the mutual information by adding skip connections between the latent layer and the layers of the decoder \citep{dieng2018avoiding}. Proposals that constrain the structure of the model do so by reducing the capacity of the decoder \citep{bowman2015generating,yang2017improved, gulrajani2016pixelvae}, by adding skip connections to each layer of the decoder \citep{dieng2018avoiding}, or by imposing constraints on encoder structure \citep{razavi2019preventing}. Taking a different approach, \citet{tomczak2017vae} and \citet{van2017neural} replace simple Gaussian priors with more complex ones such as a mixture of Gaussians.

The most recent of these proposals are $\delta$-VAE \citep{razavi2019preventing} and SKIP-VAE \citep{dieng2018avoiding}. $\delta$-VAE imposes a lower bound on KL term to prevent it from getting pushed to zero. One of the drawbacks of this approach is the fact that it introduces yet another hyper-parameter to tune carefully.Also, the model uses dropout to regularize the decoder, reducing the effective capacity of the decoder during training. It is not clear how effective the proposed method is when training more powerful decoders without such regularization. Moreover, the proposal includes an additional constraint on encoder structure, named the anti-causal encoder. 

SKIP-VAE , on the other hand, proposes to lower bound mutual information by adding skip connections from latent layers to each layer of decoder. One drawback of this approach is that it introduces additional non-linear layer per each hidden layer, resulting in more parameters to optimize. Moreover, its advantage is not clear in cases, where one can increase capacity of decoder by increasing number of units in each layer (or number of channels in CNN-based decoders) rather than adding more layers.

\section{The problem statement:}
When we train a VAE model, we ideally want to end up with a model that can reconstruct a given input well and can generate new samples in high quality. Good reconstruction requires extracting the most salient features of data and storing them on latent variable ('Encoder + Latent layer' part of the model). Generating good samples requires a generative model ('Latent layer + Decoder' part) with a model distribution that is a good approximation to actual data distribution.

However, there tends to be a trade-off between reconstruction quality of a given input, and quality of new samples. To understand why we have such a trade-off, we can start by looking at ELBO objective function\footnote{Ignoring $\mathbb{E}_{p_{d}(x)}\left[.\right]$ term for clarity}:

\begin{equation}
\begin{split}
        \mathcal{L}_{\mathrm{ELBO}} &=\mathbb{E}_{q_\phi(z \vert x)} \left[\log p_\theta(x \vert z) \right]-\mathrm{KL}(q_\phi(z\vert x) \Vert p(z)) \\
\end{split}
\end{equation} 

Maximizing this objective function increases  $p_\theta(x)$, the probability of generating each observed data x while decreasing distance between $q(z|x)$ and prior $p(z)$. Pushing $q(z|x)$ closer to $p(z)$ makes latent code less informative i.e. $z$ is influenced less by input data $x$.

The reason why the KL term can be problematic becomes more clear when we look at the KL loss term typically modelled with log of variance during optimization:

\begin{equation}\label{kl_term}
\begin{split}
        KL_{loss} =\frac{1}{2} \sum_{d=1}^{D}\left[[\mu_d^{(i)}]^2 + \left[\exp(\log\sigma^2)\right]_d^{(i)}-(\log\sigma^2)_d^{(i)}-1\right]
\end{split}
\end{equation} 

where D is the dimension of latent variable, and $i$ refers to $i^{th}$ sample. Noting that the mean is in L2 norm, minimizing the KL term leads to pushing the each dimension of the mean,$\mu_d^{(i)}$, to zero while pushing $\sigma^2$ towards 1. This makes estimated posterior less informative and less dependent on input data. The problem gets worse when dimension of latent variable, D, increases, or when the KL term is multiplied with a coefficient  $\beta>1$ \citep{higgins2017beta}. Ideally, we want to be able to distinguish between different input samples. This can be achieved by having distinctive means and variances for clusters of samples. This is where MMD might have advantage over the KL divergence. Matching distributions using MMD can match their sample means although their variance might still differ.

\section{Proposal: $\mu$-VAE}\label{proposal_uvae}
We can emulate behaviour of MMD by modifying the KL term. We do so by changing L2 norm of mean,  $\sum_{i=1}^{D}[\mu_d^{(i)}]^2$ to L1 norm, $\lvert\sum_{i=1}^{D}\mu_d^{(i)}\rvert$. Re-writing it for $B$ samples, we have:

\begin{equation}
\frac{1}{B}\lvert\sum_{i=1}^{B}\sum_{d=1}^{D}\mu_d^{(i)}\rvert
\end{equation}

It is important to note that we are taking absolute value of sum of sample means. This new formulation results in aggregate mean of samples to be zero (i.e. same mean as that of prior distribution) while allowing samples to spread out and enabling model to encode information about input data onto $z$. It should be noted that this new L1 norm of $\mu$ can push individual mean estimates to very high values if it is not constrained. To avoid that, L2 norm of means for each sample is clipped by a pre-determined value during optimization. Based on experiments, it is found that clipping L2 norm of sample means by three times square root of latent dimension works well in general although bigger values might help improve results in tasks such as classification:

\begin{equation}
\Vert\mu_{sample}\Vert \leq 3*\sqrt{z_{dim}}
\end{equation}

This method will be referred as latent clipping for the rest of this work. In addition, the remaining terms in the KL loss can be kept as is, i.e.  $\left[\exp{(\log\sigma^2)}-\log\sigma^2-1\right]$, or we can just optimize for subset of it by using either "$\log\sigma^2$", or "$\left[\exp{(\log\sigma^2)}-1\right]$" term since each method will push $\log\sigma^2$ towards zero (i.e. variance towards one). $\log\sigma^2$ is chosen in this work since it is simpler.

Finally, the $\mu$-VAE objective function can be formulated as follows:

\begin{equation}
        \boxed{
        \mathcal{L_{\mu\text{-VAE}}} =\frac{1}{B} 
        \left[ 
        \sum_{i=1}^{B}\sum_{j=1}^{J}\Vert x_j^{(i)}-x_j'^{(i)} \Vert^2 +  \lvert\sum_{i=1}^{B}\sum_{d=1}^{D}\mu_d^{(i)}\rvert  + \sum_{i=1}^{B}\sum_{d=1}^{D}\left[\log\sigma^2\right]_d^{(i)}  
        \right]
        }
\end{equation} 

where first term is reconstruction loss, $B$ refers to batch size since aggregated mean is computed over batch samples, $J$ refers to dimension of data, $D$ refers to dimension of latent variable, $x$ is original input, and $x'$ is reconstructions. 

\subsection{Latent Clipping:}
To visualize the implications of the latent clipping, a toy VAE model shown in Table~\ref{table:toy_model} in Appendix~\ref{appendix_a} is used. Figure~\ref{fig:latent_clipping_comparison} compares three cases, in which a VAE model is trained on MNIST dataset using ReLu, Tanh, and Leaky ReLu activation functions for each case, and the latent layer is visualized to observe clustering of digits. Those three cases are: i) Standard VAE objective  with the KL divergence, ii) Standard VAE objective  with the KL divergence + latent clipping, and iii) $\mu$-VAE objective function + latent clipping. Two observations can be made:
\begin{enumerate}
    \itemsep0em
    \item Latent clipping might help improve smoothness of latent space, even in the case of standard VAE objective, ELBO.
    \item $\mu$-VAE objective function seems to work well.
\end{enumerate}

\begin{figure}[!ht]
    \begin{center}
    \hbox{\hspace{3em}
    \includegraphics[width=0.25\linewidth]{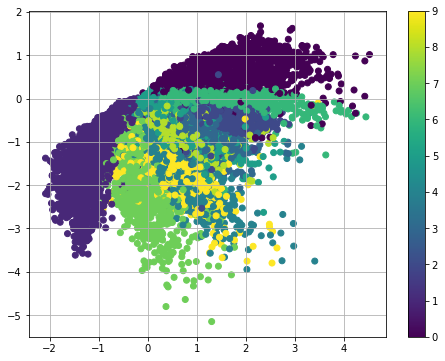}
    \includegraphics[width=0.25\linewidth]{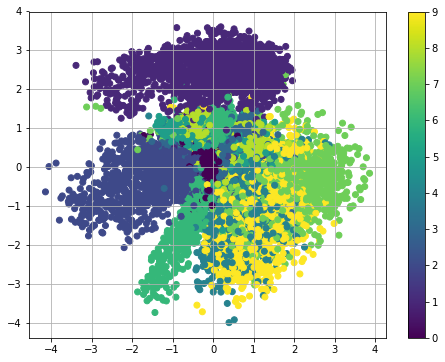}
    \includegraphics[width=0.25\linewidth]{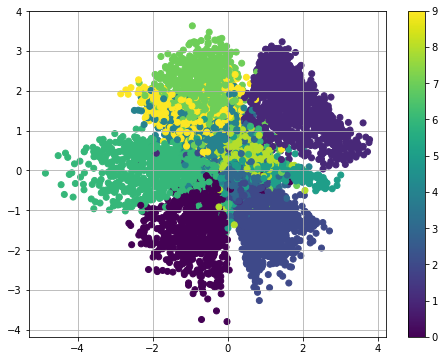}
    }
    \hbox{\hspace{3em}
    \includegraphics[width=0.25\linewidth]{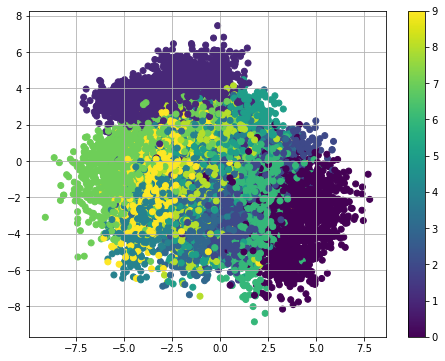}
    \includegraphics[width=0.25\linewidth]{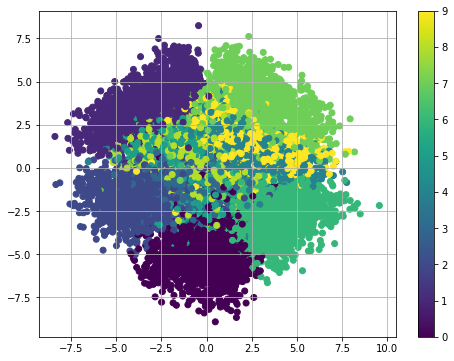}
    \includegraphics[width=0.25\linewidth]{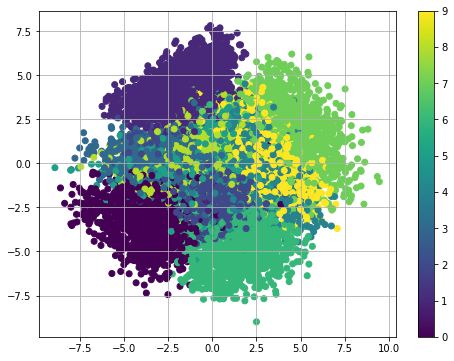}
    }    
    \hbox{\hspace{3em}
    \includegraphics[width=0.25\linewidth]{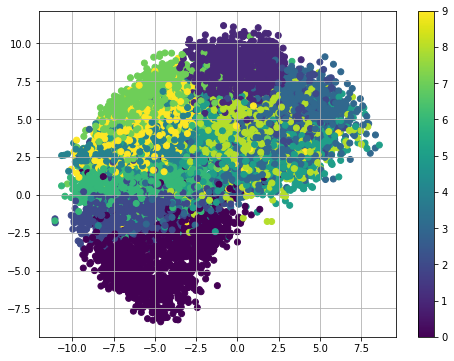}
    \includegraphics[width=0.25\linewidth]{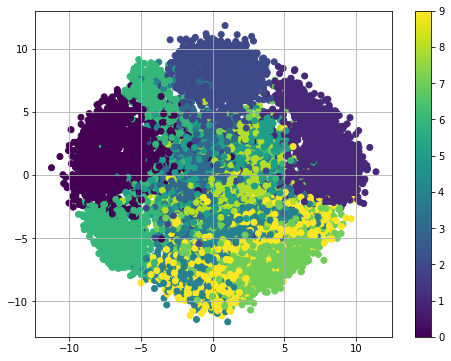}
    \includegraphics[width=0.25\linewidth]{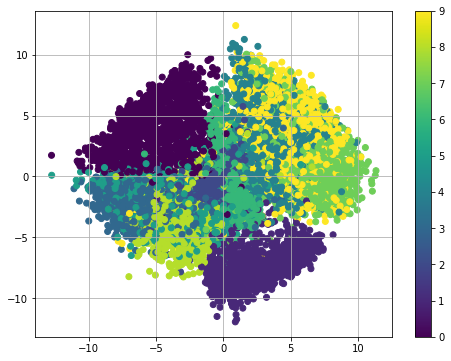}
    }        
   \caption{\textbf{Clustering of MNIST:} From left to right: Tanh, ReLu, Leaky ReLu. 
   \textbf{Top row:} Standard VAE training with ELBO objective,
   \textbf{Middle row:} Standard VAE with ELBO objective and latent clipping,
   \textbf{Bottom row:} $\mu$-VAE objective function and latent clipping.
   }
   \vspace{-1.5em}
   \label{fig:latent_clipping_comparison}
    \end{center}
\end{figure}

\section{Experiments}

To test the effectiveness of $\mu$-VAE objective, a CNN-based VAE model is designed and trained on MNIST and MNIST Fashion using same hyper-parameters for both datasets. Centered isotropic Gaussian prior, $p(z)\sim\mathcal{N}(0,\,1.0)$, is used and the true posterior is approximated as Gaussian with an approximately diagonal co-variance. No regularization methods such as dropout, or techniques such as batch-normalization is used to avoid having any extra influence on the performance, and to show the advantages of the new objective function.

The model is trained with four different objective functions: i) VAE (ELBO objective), ii) $\beta$-VAE with $\beta=4$, iii) $\mu$-VAE\#1 s.t. $\Vert\mu_{sample}\Vert \leq 3*\sqrt{z_{dim}}$ and  iv) $\mu$-VAE\#2 s.t. $\Vert\mu_{sample}\Vert \leq 6*\sqrt{z_{dim}}$, where $z_{dim}=10$.  Details of architecture, objective functions, hyper-parameters, and training are described in Appendix~\ref{appendix_b}. 

During training of the models, a simple three layer fully connected classifier is also trained over 10 dimensional latent variable to learn to classify data using features encoded on latent variable. Classifier parameters are updated when encoder and decoder parameters are frozen and vice versa so that classifier has no impact on how information is encoded on the latent variable.

\subsection{Evaluation}
Evaluation of the generative model is done qualitatively in the form of inspecting quality, and diversity of samples. Posterior collapse is assessed by comparing reconstructions of input data to observe whether the decoder ignores latent code encoded by input data and by comparing the KL divergences obtained for each model. For all three objective functions, the KL divergence is measured using standard KL formula in Equation~\ref{kl_term}. Moreover, the accuracy of the classifier trained on latent variable is used as a measure of how well the latent variable represents data \citep{dieng2018avoiding}. Higher classification accuracy reflects a better representation of data and opens doors to use latent representation for downstream tasks such as classification. 

\subsection{Results}
Figure~\ref{fig:fashion_loss_curves} shows training curves for MNIST Fashion dataset (MNIST results can be seen in Appendix~\ref{appendix_c}). The new objective function results in lower reconstruction loss, higher KL divergence, and higher classification accuracy. Higher KL divergence and classification accuracy can be interpreted as a sign of learning a more informative latent code. $\beta$-VAE performs the worst across all metrics as expected. The reason is that $\beta$ factor encourages latent code to be less informative, and is known to result in worse reconstruction quality \citep{higgins2017beta}. 

\begin{figure}[!ht]
    \begin{center}
    \hbox{\hspace{0em} 
    \includegraphics[keepaspectratio, width=0.65\paperwidth]{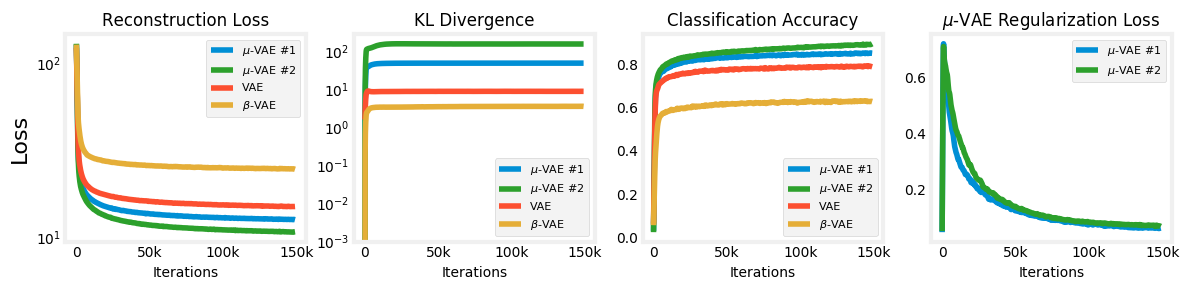} 
    }

   \caption{Training curves of the model trained on MNIST Fashion using each of four objectives: VAE (ELBO), $\beta$-VAE ($\beta=4$), $\mu$-VAE\#1 and  $\mu$-VAE\#2. Plots from left to right: reconstruction loss, KL divergence, classification accuracy and regularization loss of $\mu$-VAE ($\lvert\sum_{n=1}^{B}\mu_n\rvert  + \sum_{n=1}^{B}\left[\log\sigma^2\right]_n$) i.e. the term that replaces KL.}
   \label{fig:fashion_loss_curves}
    \end{center}
\end{figure}

\begin{figure}[!ht]
    \begin{center}
    \hbox{\hspace{0em} 
    \includegraphics[keepaspectratio, width=0.16\paperwidth]{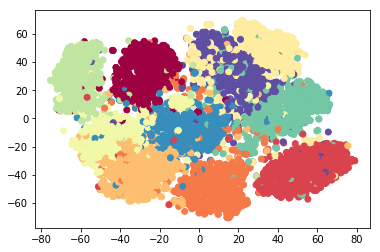} 
    \includegraphics[keepaspectratio, width=0.16\paperwidth]{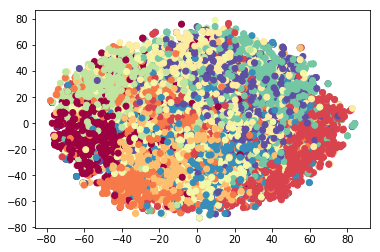} 
    \includegraphics[keepaspectratio, width=0.16\paperwidth]{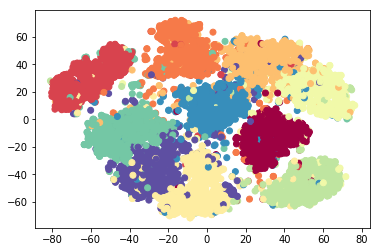} 
    \includegraphics[keepaspectratio, width=0.16\paperwidth]{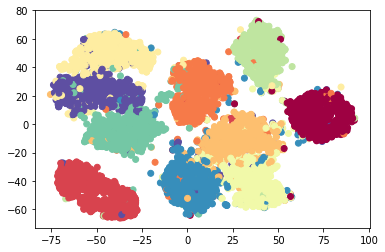} 
    }
    \hbox{\hspace{0em} 
    \includegraphics[keepaspectratio, width=0.16\paperwidth]{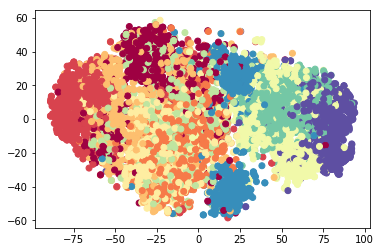} 
    \includegraphics[keepaspectratio, width=0.16\paperwidth]{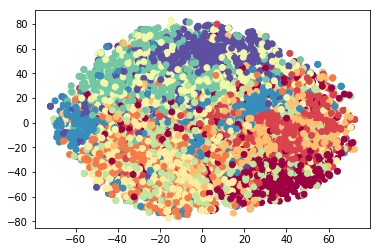} 
    \includegraphics[keepaspectratio, width=0.16\paperwidth]{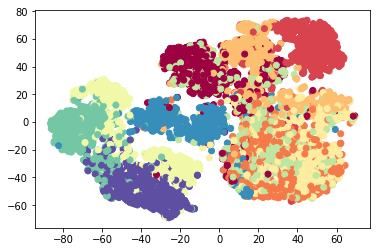} 
    \includegraphics[keepaspectratio, width=0.16\paperwidth]{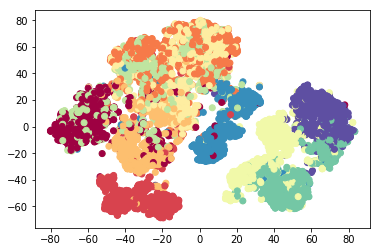} 
    }
   \caption{Clustering of samples from latent layer obtained using t-SNE and test datasets of MNIST (top row) and MNIST Fashion (bottom row). From left to right: VAE, $\beta$-VAE, $\mu$-VAE\#1, and $\mu$-VAE\#2.}
   \vspace{-2em}
   \label{fig:sample_clustering}
    \end{center}
\end{figure}

Figure~\ref{fig:sample_clustering} shows results of t-SNE \citep{maaten2008visualizing} of samples obtained using test data for both datasets. VAE seems to able to distinguish all ten digits, but performs worse in MNIST Fashion. $\beta$-VAE pushes samples closer as expected, which explains why its performance is low in classification task. $\mu$-VAE, on the other hand, is able to cluster similar samples together in both datasets. Moreover, when upper-bound on $\Vert\mu_{sample}\Vert$ is increased, it spreads out clusters of samples, making them easier to distinguish. Hence, upper-bound used in latent clipping can be a knob to control distance between samples. Also, we should note that we can achieve similar clustering results to the one obtained by VAE through reducing the upper-bound. Smaller upper-bound closes gaps between the clusters.

Table~\ref{table:test_accuracy} lists classification accuracy obtained using test datasets. $\mu$-VAE performs the best as expected since it is able to push the clusters apart. Higher upper bound on  $\Vert\mu_{sample}\Vert$ results in a higher classification accuracy. Also, it should be noted that reported accuracy numbers can be improved, but the purpose of this test was to show that new objective function can reliably be used in downstream tasks such as classification.

\begin{table}[ht]
\centering
\caption{ Comparing test accuracy of classifiers trained on latent variables of each model.}
    \begin{tabular}[t]{ccc}
        \hline
        \textbf{Model}&\multicolumn{1}{l}{\textbf{MNIST}}&\multicolumn{1}{l}{\textbf{MNIST Fashion}}\\
        \hline
        \hline

        \textbf{VAE}         & 93.422           &  78.01  \\
        \textbf{$\beta$-VAE} & 62.45            &  62.35    \\
        \textbf{$\mu$-VAE \#1}& \textbf{95.28}  &  \textbf{82.71}    \\
        \textbf{$\mu$-VAE \#2}& \textbf{96.44}  &  \textbf{84.260}    \\

        \hline
        \hline

    \end{tabular}
    \label{table:test_accuracy}
\end{table}%



\begin{figure}[!ht]
    \begin{center}
    \hbox{\hspace{2em} 
    \includegraphics[keepaspectratio, width=0.55\paperwidth]{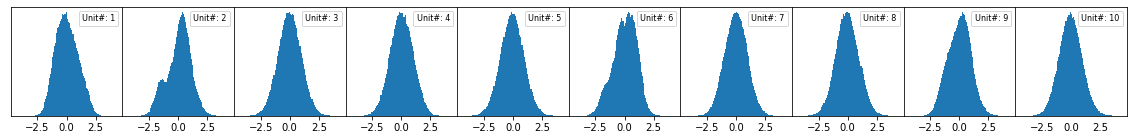} 
    }
    \hbox{\hspace{2em} 
    \includegraphics[keepaspectratio, width=0.55\paperwidth]{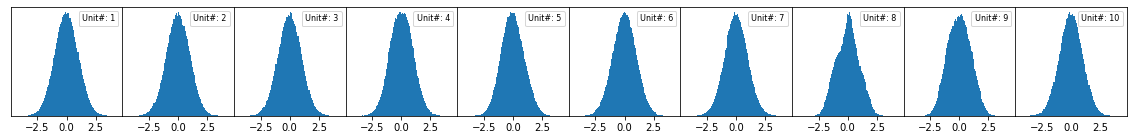} 
    }
    \hbox{\hspace{2em} 
    \includegraphics[keepaspectratio, width=0.55\paperwidth]{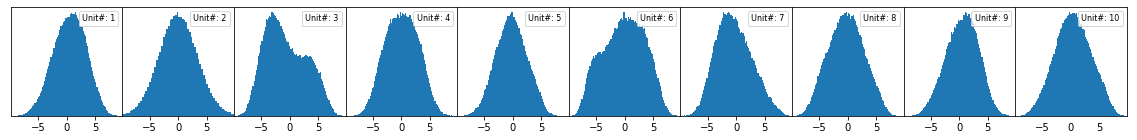}
    }
    \hbox{\hspace{2em} 
    \includegraphics[keepaspectratio, width=0.55\paperwidth]{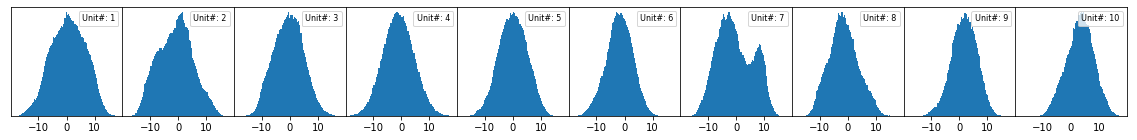} 
    }
    
   \caption{Sample distribution obtained at each dimension of latent variable using test dataset of MNIST Fashion. From top to bottom: VAE, $\beta$-VAE, $\mu$-VAE\#1, and $\mu$-VAE\#2.}
   \vspace{-1em}
   \label{fig:sample_distributions}
    \end{center}
\end{figure}

Figure~\ref{fig:sample_distributions} compares sample distributions obtained at each dimension of latent variable using test dataset of MNIST Fashion for each objective function. $\beta$-VAE samples follow $N(0,1)$ prior very closely, and hence resulting in the smallest KL divergence term. Sample distributions from the most dimensions of VAE are also close to prior distribution although some of them show a multi-modal behavior. Sample distributions from both $\mu$-VAE\#1 \& \#2 result in zero mean, but they are more spread out as expected. Spread is controlled by upper-bound on $\Vert\mu_{sample}\Vert$. Similar to VAE, some sample distributions show a multi-modal behavior.

\begin{figure}[!ht]
    \begin{center}
    \hbox{\hspace{0em} 
    \includegraphics[keepaspectratio, width=0.15\paperwidth]{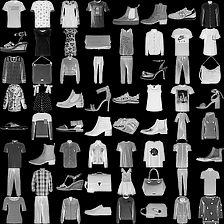} 
    \includegraphics[keepaspectratio, width=0.15\paperwidth]{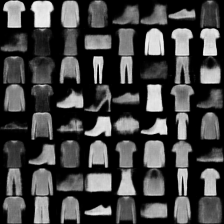} 
    \hspace{0.5em} 
    \includegraphics[keepaspectratio, width=0.15\paperwidth]{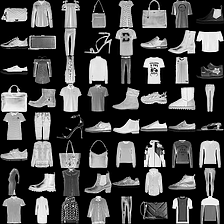} 
    \includegraphics[keepaspectratio, width=0.15\paperwidth]{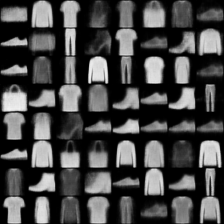}     
    }
    \vspace{0.5em} 

    \hbox{\hspace{0em} 
    \includegraphics[keepaspectratio, width=0.15\paperwidth]{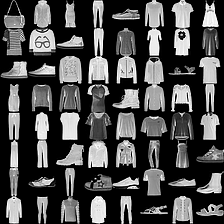} 
    \includegraphics[keepaspectratio, width=0.15\paperwidth]{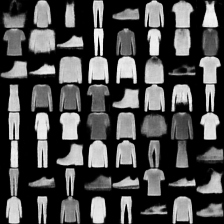} 
    \hspace{0.5em} 
    \includegraphics[keepaspectratio, width=0.15\paperwidth]{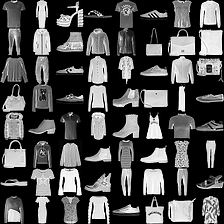} 
    \includegraphics[keepaspectratio, width=0.15\paperwidth]{images/latest_results/tvae_cn3_test_fashion.png}     
    
    }
    
   \caption{Reconstructions of input data obtained using test dataset of MNIST Fashion. Top row: VAE (left), $\beta$-VAE (right). Bottom row: $\mu$-VAE\#1 (left), $\mu$-VAE\#2 (right).}
   \vspace{-1em}
   \label{fig:input_reconstructions_test_data}
    \end{center}
\end{figure}

Figure~\ref{fig:input_reconstructions_test_data} shows reconstruction of input using test dataset. $\beta$-VAE reconstructions are either blurrier, or wrong, the latter of which is a sign of posterior collapse. VAE performs better, and both versions of $\mu$-VAE gives the best reconstruction quality. 

\begin{figure}[!ht]
    \begin{center}
    
    \hbox{\hspace{-0.5em} 
    \includegraphics[keepaspectratio, width=0.16\paperwidth]{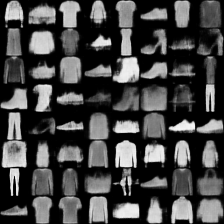} 
    \includegraphics[keepaspectratio, width=0.16\paperwidth]{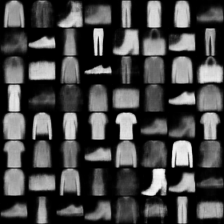} 
    \includegraphics[keepaspectratio, width=0.16\paperwidth]{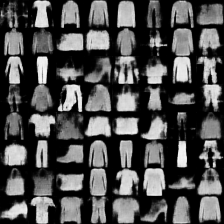}
    \includegraphics[keepaspectratio, width=0.16\paperwidth]{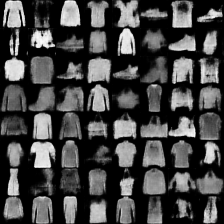}     
    }

   \caption{Random samples drawn from multi-variate Gaussian, N(0, $\sigma$).From left to right, model ($\sigma$): VAE ($\sigma$=1), $\beta$-VAE ($\sigma$=1), $\mu$-VAE\#1 ($\sigma$=3), and $\mu$-VAE\#2 ($\sigma$=3). Higher $\sigma$ is used for $\mu$-VAE models since their samples are more spread out.}
   \label{fig:random_samples}
    \end{center}
\end{figure}

Figure~\ref{fig:random_samples} shows images generated using random samples drawn from multivariate Gaussian, N(0, $\sigma$), where $\sigma=1$ is for VAE and $\beta$-VAE while it is 3 for $\mu$-VAE since their samples are more spread out (MNIST results can be seen in Appendix~\ref{appendix_c}). We can observe that some samples generated from $\mu$-VAE models have dark spots. This is because the model is trying to generate texture on these samples. This can also be observed in samples of VAE model, but it is less pronounced. However, samples from $\beta$-VAE do not show any such phenomena since the model perhaps learns global structure of shapes while ignoring local features. Failing to capture local structures is a known problem in latent variable models \citep{larsen2015autoencoding, razavi2019preventing}.

\begin{figure}[!ht]
    \begin{center}
    \hbox{\hspace{0em} 
    \includegraphics[keepaspectratio, width=0.6\paperwidth]{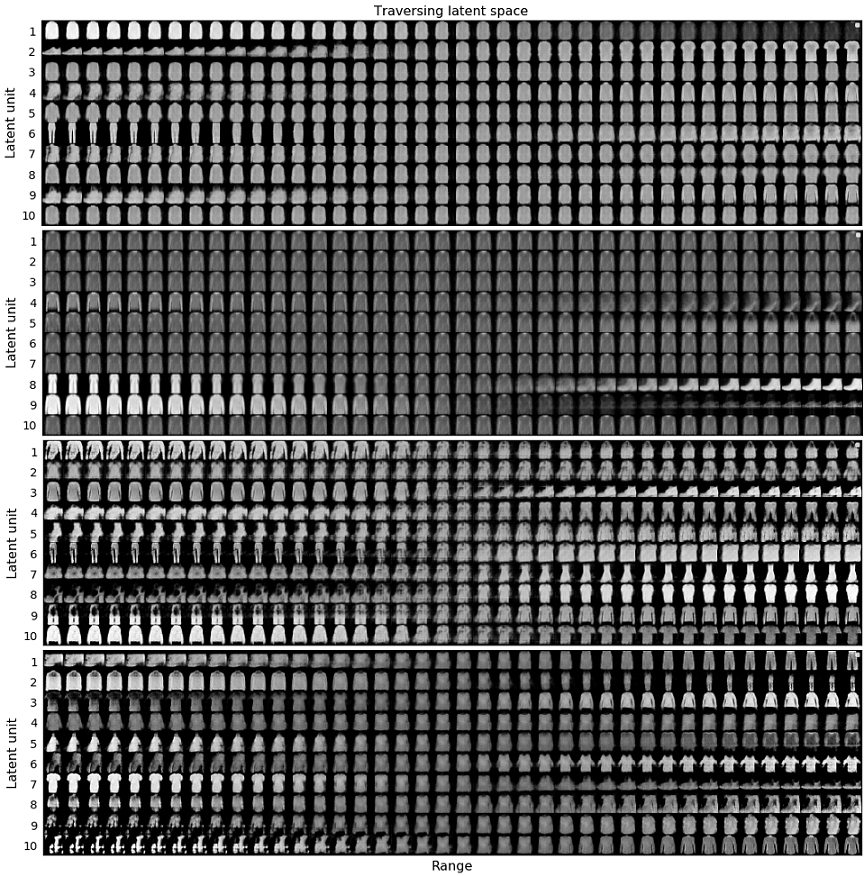} 
    }

   \caption{Traversing each dimension of latent variable in 40 steps. From top to bottom, the model and sweeping range used for the model: VAE [-2,2], $\beta$-VAE  [-2,2], $\mu$-VAE\#1  [-10,10], $\mu$-VAE\#2  [-20,20].}
   \label{fig:fashion_latent_traverse_small}
    \end{center}
\end{figure}

Figure~\ref{fig:fashion_latent_traverse_small} shows latent traverse in each dimension of latent variable for MNIST Fashion (MNIST results can be seen in Appendix~\ref{appendix_c}). Each dimension of VAE and $\beta$-VAE is swept in $[-2,2]$ range linearly while other dimensions are kept at zero. For $\mu$-VAE models, ranges of $[-10,10]$ and $[-20,20]$ are used since samples are more spread out. $\beta$-VAE gives mostly similar classes of objects, a sign that most dimensions of latent variable are not very informative. VAE is slightly better. However, both $\mu$-VAE models learn diverse classes of objects across different dimensions. Moreover, they learn different classes on opposite sides of same dimension. This is encouraging since it shows its power to learn rich representations. 

\section{Summary}
In this work, a new objective function is proposed to mitigate posterior collapse observed in VAEs. It is shown to give better reconstruction quality and to learn good representations of data in the form of more informative latent codes. A method, named latent clipping, is introduced as a knob to control distance between samples. Samples can be pushed apart for tasks such as classification, or brought closer for smoother transition between different clusters of samples. Unlike prior work, the proposed method is robust to parameter tuning, and does not constraint encoder, or decoder structure. It can be used as a direct replacement for ELBO objective. Moreover, the proposed method is demonstrated to learn representations of data that can work well in downstream tasks such as classification. Applications of $\mu$-VAE objective with more powerful decoders in various settings can be considered as a future work.


\bibliography{iclr2020_conference}
\bibliographystyle{iclr2020_conference}

\clearpage
\appendix
\section{Model architecture used for toy example of latent clipping}\label{appendix_a}
\begin{table}[ht]
\centering
\caption{ Toy VAE Model for visualization of experiments}
    \hspace*{-0.5cm}\begin{tabular}[t]{ll}
        \hline
        \textbf{Function}&\multicolumn{1}{c}{\textbf{Layer}}\\
        \hline
        \hline

        & Fully Connected Layer 784x512 + Activation  \\
        \textbf{Encoder}& Fully Connected Layer 512x512 + Activation      \\
        & Fully Connected Layer 512x512 + Activation     \\

        \hline
        
        \textbf{Sampling}& Linear layer with 2 units \\
        
        \hline

        & Fully Connected Layer   2x512 + Activation   \\
        \textbf{Decoder}& Fully Connected Layer 512x512 + Activation \\
        & Fully Connected Layer 512x512 + Activation \\
        & Fully Connected Layer 512x784 + Sigmoid \\

        \hline
        \hline
        &*Activation functions are all either Leaky ReLu (alpha=0.2), \\
        &ReLu, or Tanh  as part of the experimentation. \\
        \hline

    \end{tabular}
    \label{table:toy_model}
\end{table}%

\section{Model architecture, objective functions and details of training.}\label{appendix_b}

\textbf{Optimization:} 
In all experiments, learning rate of 1e-4 and batch size of 64 are used. Adam algorithm with high momentum ($\beta1 = 0.9, \beta2=0.999$) is used as optimizer. High momentum is chosen mainly to let most of previous training samples influence the current update step. 

For reconstruction loss, mean square error, $\Vert x-x' \Vert^2$, is used for all cases. 
As for initialization, since the model consists of convolutional layers with Leaky ReLu in both encoder and decoder, Xavier initialization is used \cite{glorot2010understanding}. Thus, initial weights are drawn from a Gaussian distribution with standard deviation (stdev) of $\sqrt{2/N}$, where N is number of nodes from previous layer. For example, for a kernel size of 3x3 with 32 channels, N = 288, which results in stdev of $0.083$.


Objective functions are shown in Table~\ref{table:objective_for_3D}, where $\mu$-VAE objective is written explicitly to avoid any ambiguity in terms of how batch statistics are computed. 

Table~\ref{table:model_architecture} shows model architecture as well as classifier used for all experiments. It consists of CNN-based encoder and decoder while classifier is three layer fully connected neural network. They all use Leaky Relu activation and learning rate of 1e-4.

\begin{table}[ht]
\centering
\caption{Objective functions\protect\footnotemark}
    \hspace*{0cm}\begin{tabular}[t]{ll}

        \hline
        \hline
        
        \textbf{ELBO}&$\mathcal{L}_{e} = \mathbb{E}_{q_\phi(z \vert x)} \left[\log p_\theta(x \vert z) \right] -  \mathrm{KL}(q_\phi(z\vert x) \Vert p(z))$  \\
        
        \textbf{$\beta$-VAE}&$\mathcal{L}_{\beta} = \mathbb{E}_{q_\phi(z \vert x)} \left[\log p_\theta(x \vert z) \right] -  \beta * \mathrm{KL}(q_\phi(z\vert x) \Vert p(z))$ \\

        \textbf{$\mu$-VAE}&$\mathcal{L}_\mathrm{\mu}=\mathbb{E}_{q_\phi(z \vert x)} \left[\log p_\theta(x \vert z) \right] - \frac{1}{B} 
        \left[ \lvert\sum_{i=1}^{B}\sum_{d=1}^{D}\mu_d^{(i)}\rvert  + \sum_{i=1}^{B}\sum_{d=1}^{D}\left[\log\sigma^2\right]_d^{(i)}  
        \right]$ \\
        \hline

    \end{tabular}
    \label{table:objective_for_3D}
\end{table}%

\footnotetext{Note that each $\mathbb{E}_{q_\phi(z \vert x)}[.]$ and KL[.] is computed on expectation, $\mathbb{E}_{p_{data}(x)}\left[.\right]$, but it is not shown explicitly in the formulas above to make them more readable. Also, regularization term in $\mu$-VAE is shown explicitly to emphasize how sample means are computed over batches.}


\begin{table}[ht]
\centering
\caption{Model used for comparing VAE, $\beta$-VAE and $\mu$-VAE}
    \begin{tabular}[t]{ll}
        \hline
        \textbf{Function}&\multicolumn{1}{c}{\textbf{Layer}}\\
        \hline
        \hline
        \textbf{Encoder}& 2D Conv 28x28x64  + Leaky Relu \\
        & 2D Conv 14x14x64  + Leaky Relu \\
        & 2D Conv 7x7x64 + Leaky Relu   \\
        & Linear layer with (7x7x64)x10  \\
        \hline
        
        \textbf{Sampling}& 10 Latent layer units \\
        
        \hline
        \textbf{Decoder}& Linear Layer 10x(7x7x64) + Leaky Relu  \\
        & 2D DeConv 7x7x64 + Leaky Relu  \\
        & 2D DeConv 14x14x32 + Leaky Relu    \\
        & 2D DeConv 28x28x1 + Sigmoid  \\

        \hline

        \hline
        \textbf{Classifier}&Dense 10x1024 + Leaky Relu  \\
        &Dense 1024x1024 + Leaky Relu  \\
        &Dense 1024x1024 + Leaky Relu\\
         &Dense 1024x10 + Softmax \\

        \hline
        \hline
        &*All Leaky ReLu layers use alpha=0.2. \\
        \hline

    \end{tabular}
    \label{table:model_architecture}
\end{table}%



\clearpage
\section{MNIST Results}\label{appendix_c}

\begin{figure}[!ht]
    \begin{center}
    \hbox{\hspace{0em} 
    \includegraphics[keepaspectratio, width=0.6\paperwidth]{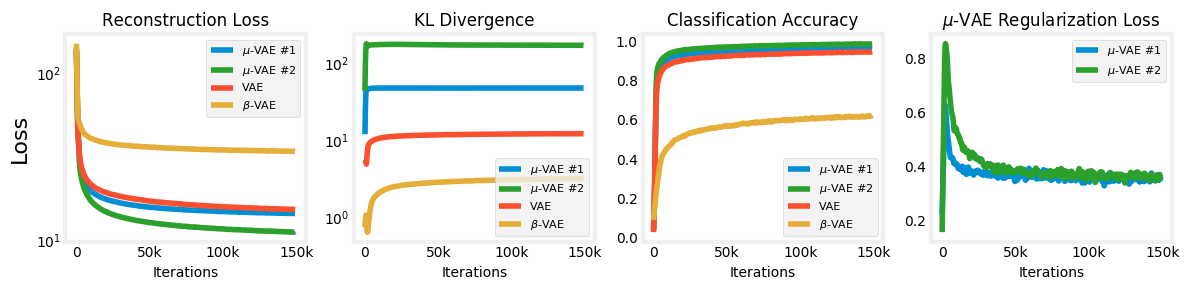} 
    }

   \caption{Training curves of the model trained on MNIST. Regularization loss of $\mu$-VAE defined as $\lvert\sum_{n=1}^{B}\mu_n\rvert  + \sum_{n=1}^{B}\left[\log\sigma^2\right]_n$, i.e. term that replaces KL.}
   \vspace{-2em}
   \label{fig:mnist_training_curve}
    \end{center}
\end{figure}

\begin{figure}[!ht]
    \begin{center}
    \hbox{\hspace{0em} 
    \includegraphics[keepaspectratio, width=0.6\paperwidth]{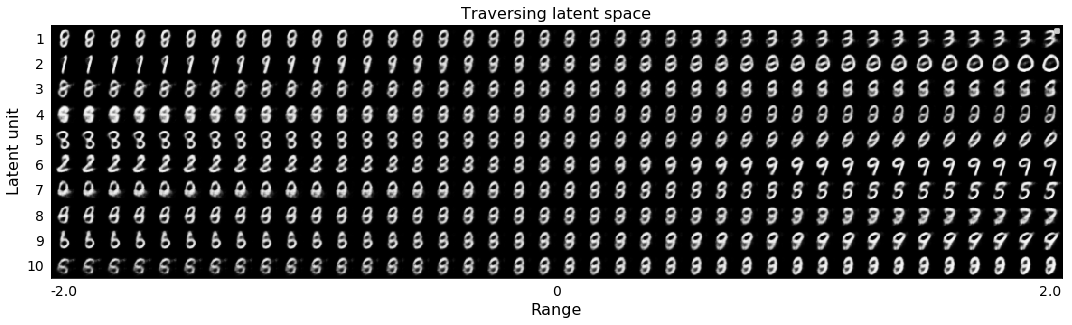} 
    }
    \hbox{\hspace{0em} 
    \includegraphics[keepaspectratio, width=0.6\paperwidth]{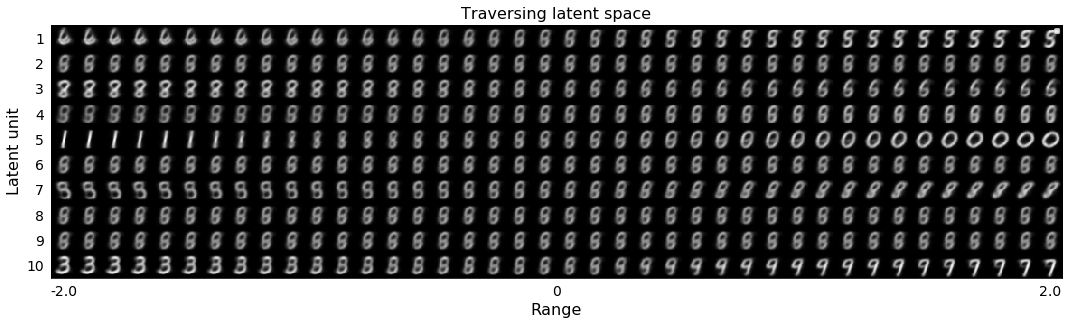} 
    }
    \hbox{\hspace{0em} 
    \includegraphics[keepaspectratio, width=0.6\paperwidth]{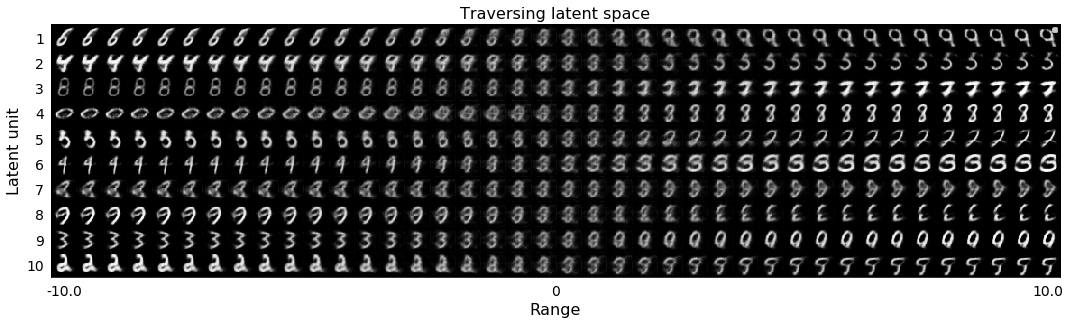}
    }
    \hbox{\hspace{0em} 
    \includegraphics[keepaspectratio, width=0.6\paperwidth]{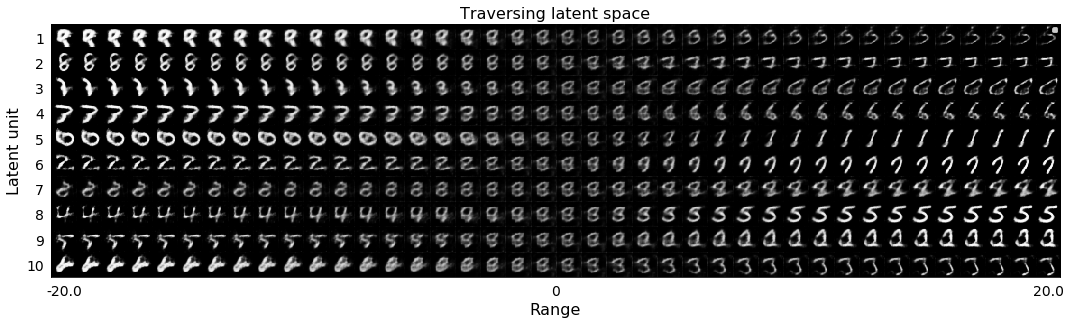} 
    }
   
   \caption{Traversing each dimension of latent variable in 40 steps. From top to bottom, model [range]: VAE [-2,2], $\beta$-VAE  [-2,2], $\mu$-VAE\#1  [-10,10], and $\mu$-VAE\#2  [-20,20].}
   \vspace{-2em}
   \label{fig:mnist_latent_traverse}
    \end{center}
\end{figure}

\begin{figure}[!ht]
    \begin{center}
    
    \hbox{\hspace{0em} 
    \includegraphics[keepaspectratio, width=0.15\paperwidth]{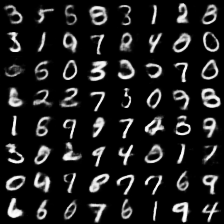} 
    \includegraphics[keepaspectratio, width=0.15\paperwidth]{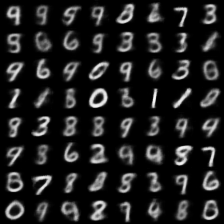} 
    \includegraphics[keepaspectratio, width=0.15\paperwidth]{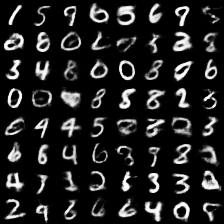}
    \includegraphics[keepaspectratio, width=0.15\paperwidth]{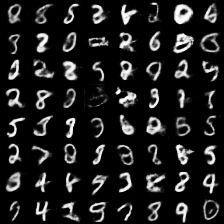}     
    }

   \caption{Random samples drawn from multi-variate Gaussian, N(0, $\sigma$).From left to right, model ($\sigma$): VAE ($\sigma$=1), $\beta$-VAE ($\sigma$=1), $\mu$-VAE\#1 ($\sigma$=3), and $\mu$-VAE\#2 ($\sigma$=3). Higher $\sigma$ is used for $\mu$-VAE models since their samples are more spread out.}
   \vspace{-2em}
   \label{fig:mnist_random_samples}
    \end{center}
\end{figure}

\begin{figure}[!ht]
    \begin{center}
    \hbox{\hspace{0em} 
    \includegraphics[keepaspectratio, width=0.65\paperwidth]{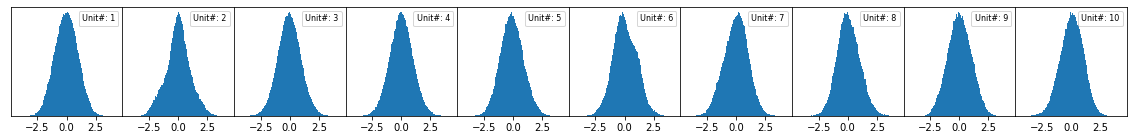} 
    }
    \hbox{\hspace{0em} 
    \includegraphics[keepaspectratio, width=0.65\paperwidth]{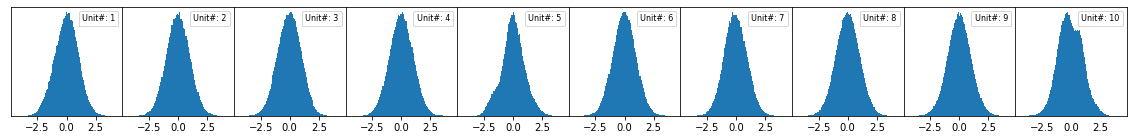} 
    }
    \hbox{\hspace{0em} 
    \includegraphics[keepaspectratio, width=0.65\paperwidth]{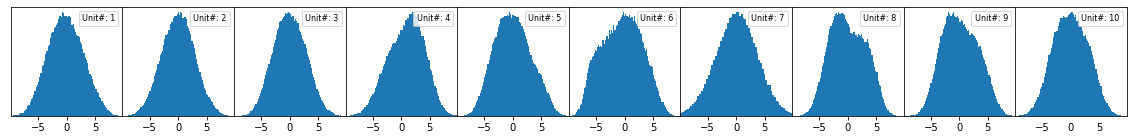}
    }
    \hbox{\hspace{0em} 
    \includegraphics[keepaspectratio, width=0.65\paperwidth]{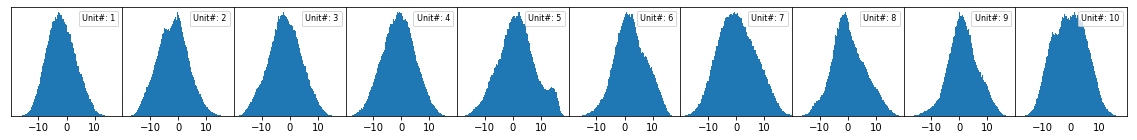} 
    }
    
   \caption{Sample distribution obtained at each dimension of latent variable using test dataset of MNIST. From top to bottom: VAE, $\beta$-VAE, $\mu$-VAE\#1, and $\mu$-VAE\#2.}
   \vspace{-2em}
   \label{fig:mnist_sample_distributions}
    \end{center}
\end{figure}

\end{document}